\documentclass[lettersize,journal]{IEEEtran}
\usepackage{cite}
\usepackage{amsmath,amssymb,amsfonts}
\usepackage{algorithmic}
\usepackage{graphicx}
\usepackage{textcomp}
\usepackage{xcolor}
\usepackage{subcaption}
\usepackage{multirow}
\usepackage{booktabs}
\usepackage{tabularx}
\usepackage{makecell}
\usepackage{url}
\usepackage{adjustbox}
\usepackage{float}
\usepackage{placeins}
\usepackage{booktabs}
\usepackage{longtable}
\usepackage{array}
\usepackage{lscape}
\usepackage[table]{xcolor}
\usepackage{caption}
\usepackage{ragged2e}

\definecolor{cluster1}{HTML}{1F77B4}
\definecolor{cluster2}{HTML}{AEC7E8}
\definecolor{cluster3}{HTML}{FF7F0E}
\definecolor{cluster4}{HTML}{FFBB78}
\definecolor{cluster5}{HTML}{2CA02C}
\definecolor{cluster6}{HTML}{98DF8A}
\definecolor{cluster7}{HTML}{D62728}
\definecolor{cluster8}{HTML}{FF9896}
\definecolor{cluster9}{HTML}{9467BD}
\definecolor{cluster10}{HTML}{C5B0D5}
\definecolor{cluster11}{HTML}{8C564B}
\definecolor{cluster12}{HTML}{C49C94}

\newcommand{\clusterlabel}[1]{%
  \mbox{%
    \raisebox{-1.3ex}{\textcolor{cluster#1}{\Large\textbullet}}%
    \,\textbf{C#1}%
  }%
}

\newcolumntype{L}[1]{
                    >{\RaggedRight\arraybackslash}p{#1}
                }
\begin{document}

\title{Beyond Triplet Plausibility: Relation Set Completion in Knowledge Graphs}

\author{Zihao~Zheng,
Borui~Cai,
Yao~Zhao,
Xin~Han,
Mengqi~Ji
        % <-this % stops a space
\thanks{Z. Zheng and X. Han are with the School of Information Technology, Deakin University, VIC 3125, Australia.
E-mail: z.zheng@deakin.edu.au; xinhan@ieee.org.}

\thanks{B. Cai is with Hangzhou International Innovation Institute, Beihang University,
Hangzhou 311115, China. E-mail: caibr@buaa.edu.cn.}

\thanks{Y. Zhao is with the School of Information Technology, Victoria University, VIC 3011, Australia. E-mail: yao.zhao@vu.edu.au.}

\thanks{M. Ji is with School of Artificial Intelligence, Beihang University, Beijing 100191, China. E-mail: jimengqi@buaa.edu.cn.}

\thanks{Corresponding author: Borui Cai}}

% The paper headers
\markboth{Journal of \LaTeX\ Class Files,~Vol.~14, No.~8, August~2021}%
{Zheng \MakeLowercase{\textit{et al.}}: Beyond Triplet Plausibility: Relation Set Completion in Knowledge Graphs}

\maketitle

\begin{abstract}
Knowledge graphs (KGs) organize real-world knowledge as triplets and underpin many downstream applications. 
Due to their inherent incompleteness, knowledge graph completion (KGC) is widely studied and is typically formulated as triplet prediction, with link prediction as the dominant paradigm. However, this formulation focuses on the incompleteness of triplet-wise information and overlooks the incompleteness of entity-relation compatibility information.
To address this limitation, we introduce a relation set completion task (RSC), which complements the link prediction task and aims to reason about missing relations that are semantically compatible with a given entity.
We further propose a Relation Set Embedding model (RelSetE), which models latent patterns among the observed relations of entities to infer missing ones.
To evaluate RelSetE, we derive three benchmark datasets from standard KG benchmarks.
Extensive experiments demonstrate that RelSetE effectively captures entity-relation compatibility patterns and performs favorably in inferring missing relations of entities.
Code and data are publicly available.\footnote{\url{https://github.com/zihao-johnson/knowledge-graph-relation-set-completion}.}
\end{abstract}


\section{Introduction}
\IEEEPARstart{K}{nowledge} graphs (KGs) organize structured real-world knowledge as triplets of the form 
$\langle head$ $entity, relation, tail$ $entity \rangle$, 
and underpin a wide range of downstream applications such as information retrieval \cite{xiong2017explicit,yang2020biomedical,yang2022development} and recommendation systems \cite{gao2023enhanced,shokrzadeh2024knowledge}.
Despite their practical importance, real-world KGs are inherently incomplete, prompting extensive research on knowledge graph completion (KGC) \cite{bordes2013translating}.
In recent years, KGC has predominantly been formulated as a link prediction (LP) task, aiming to infer missing facts by predicting the missing entities or relations in incomplete triplets. Such a formulation primarily relies on triplet-wise information, i.e., the likelihood that a given head entity, relation, and tail entity form a valid fact.
While existing LP methods focus primarily on triplet plausibility, entity–relation compatibility remains unexplored.

%%-----------------------------------------
%% entity-centric information incomplete  |
%%-----------------------------------------

In many real-world settings, KGs are constructed alongside pre-defined ontologies or domain knowledge that explicitly specify which relations are semantically compatible with different types of entities. Under this assumption, entity–relation compatibility is regarded as prior knowledge, and the set of relations compatible with each entity is assumed to be known. However, with large-scale information extraction and the emergence of open-world settings \cite{construct}, KGs continuously grow and evolve, inevitably outpacing the coverage of any pre-defined ontologies or domain knowledge, making this assumption increasingly unrealistic. As a result, the compatible relation sets of entities can no longer be reliably derived from pre-defined ontologies, as entities may either fail to fit into pre-defined ontological categories or participate in relations beyond their coverage.

This limitation is further reflected at the data level. Because KGs are inherently incomplete, the observed triplets associated with an entity usually reveal only a partial view of its full compatible relation set. In other words, the absence of a relation from the observed KG does not necessarily mean that the relation is semantically incompatible with the entity. Without recovering such missing compatibility information, the semantic representation of an entity may remain incomplete, which can further affect downstream KG-based tasks.

This highlights the need to study the completion of entity–relation compatibility information, which to the best of our knowledge remains unexplored. 
To address this gap, we introduce Relation Set Completion (RSC), a novel KGC task that operates at the entity level rather than the triplet level. Formally, given an entity, RSC aims to infer the set of relations that are semantically compatible with that entity but have not yet been associated with it in the KG.
While LP determines whether a specific triplet is plausible, RSC determines which relations are semantically applicable to an entity, making the two tasks complementary perspectives on KG completion.

%  --------------------figure 1 LP VS RSC--------
\begin{figure*}[!htbp]
    \centering
    \includegraphics[width=0.95\linewidth]{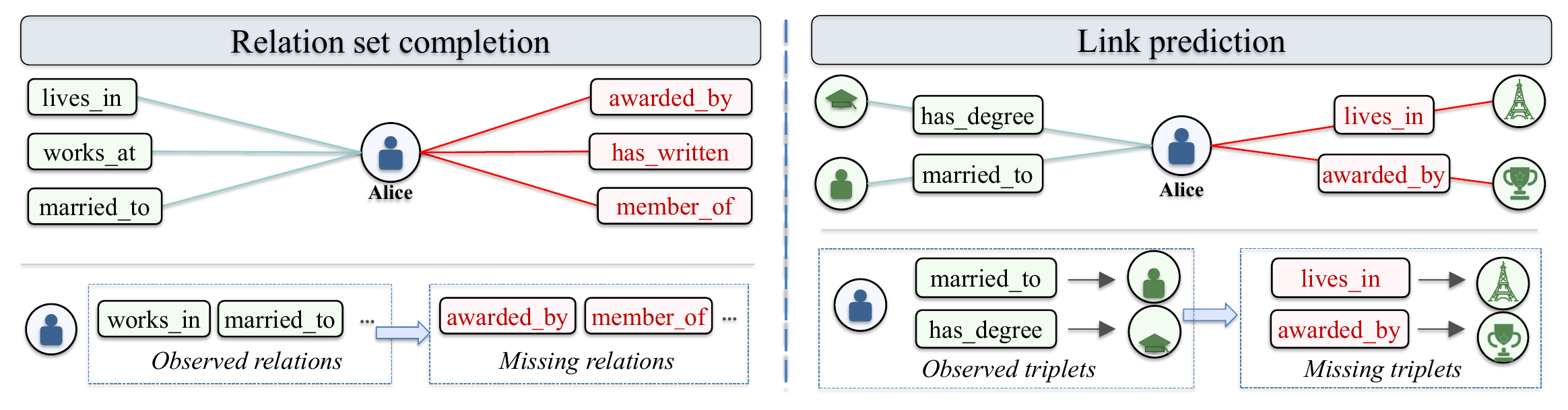}
    \caption{Comparison of link prediction and the proposed relation set completion.
    }
    \label{fig:lp_vs_rsc}
\end{figure*}
%  --------------------figure 1 END---------------

%%---------------
%% RelSetE      |
%%---------------
We then propose Relation Set Embedding (RelSetE), a simple yet effective baseline designed to establish an initial benchmark for the proposed RSC task. RelSetE formulates RSC as predicting missing compatible relations conditioned on an entity’s observed relation set. Since relation sets are naturally permutation-invariant, variable-sized, and free of duplicates, RelSetE adopts a set-aware modeling approach. Specifically, it encodes the observed relation set into a hidden representation using multi-head attention and attention pooling, and predicts missing compatible relations by measuring the similarity between this representation and candidate relation embeddings. We further construct three datasets tailored for RSC from standard KGC benchmarks and conduct comprehensive experiments. The results show that RelSetE consistently outperforms strong baselines in prediction accuracy.
Our main contributions are summarized as follows:
\begin{itemize}
    % \begin{enumerate}
        \item 
        We introduce Relation Set Completion (RSC), a novel entity-level KGC task that targets the completion of entity–relation compatibility information, complementing traditional triplet-level link prediction.
        \item 
        We propose Relation Set Embedding (RelSetE), a set-aware model that captures semantic dependencies among relations to predict missing compatible relations for an entity.   
        \item 
        We construct three benchmark datasets for RSC from standard KGC datasets and conduct comprehensive experiments, demonstrating that RelSetE consistently outperforms strong baselines and effectively predicts missing compatible relations across datasets.
\end{itemize}

%  --------------------figure 2 RelSetE architectures
\begin{figure*}[!htbp]
    \centering
    \includegraphics[width=\linewidth]{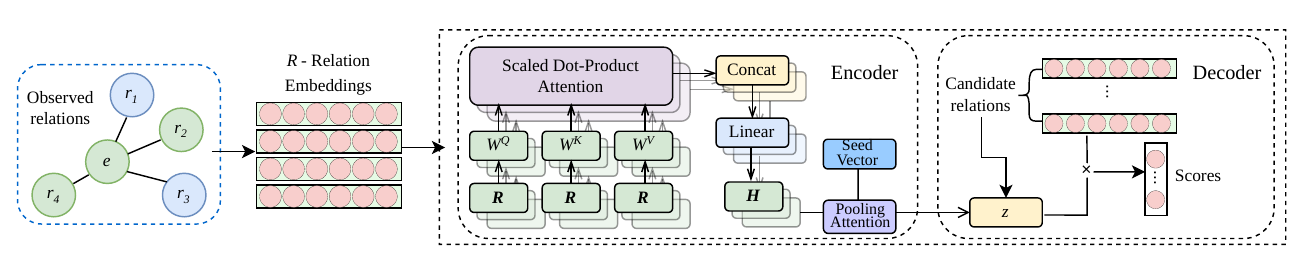}
    \caption{The overall framework of RelSetE, which predicts missing compatible relations from an entity’s observed relations using relation embeddings, a permutation-invariant encoder, and a compatibility scoring decoder.
    }
    \label{fig:RelSetE}
\end{figure*}
%  ----------------------------------------

\section{Related Work}
\label{sec:related_work}

% ----------------
\subsection{Knowledge Graph Completion}

Knowledge graphs (KGs) encode real-world facts as triplets and support a wide range of downstream applications, yet they are inherently incomplete. To mitigate this, knowledge graph completion (KGC) has been widely studied, with link prediction (LP), which assesses the plausibility of candidate triplets, being the predominant task.
Existing LP approaches mainly fall into two broad categories. The first is probabilistic modeling, including Markov logic networks \cite{lu2017link}, which integrate first-order logic with probabilistic graphical models to handle uncertain relational dependencies, and Bayesian networks \cite{lao2010relational, zhai2018method}, which capture probabilistic dependencies among entities and relations for relational inference.
The second, and more dominant, category is representation learning, which embeds entities and relations into continuous vector spaces. Within this paradigm, translational models \cite{bordes2013translating, wang2014knowledge, chang2017knowledge} interpret relations as translations between head and tail entity embeddings; factorization-based models \cite{yang2014embedding, trouillon2016complex, balavzevic2019tucker, zhang2020learning} decompose the KG tensor or scoring function to capture latent relational structure; and neural architectures \cite{socher2013reasoning, dettmers2018convolutional} learn nonlinear scoring functions for triplet plausibility.
Beyond these foundations, several lines of work extend standard LP with auxiliary information, such as path-aware reasoning \cite{lin2015modeling}, which exploits multi-hop relational paths as contextual evidence, and knowledge-aware reasoning \cite{ma2023kr}, which incorporates external structured knowledge to enhance inference.
Different from these works, this paper introduces RSC, a novel entity-level KGC task that shifts the focus from triplet plausibility to entity–relation compatibility, aiming to complete the full set of relations that are semantically compatible with each entity.

% -------------
\subsection{Set Modeling}
Set modeling studies how to learn representations or make predictions from data whose elements form an unordered collection.
Early work \cite{vinyals2015order} investigates the role of ordering in sequence-to-sequence models for set-structured problems, proposing a read-process-write framework that encodes the input set in a permutation-invariant manner while generating the output auto-regressively. A parallel line of research focuses on learning permutation-invariant representations from set-structured inputs. Bayesian Sets \cite{ghahramani2005bayesian} provide an early probabilistic formulation for modeling exchangeable sets, but their applicability to large-scale data is limited. DeepSet \cite{zaheer2017deep} provides a formal characterization of permutation-invariant functions, showing that any such function can be decomposed into a shared element-wise transformation followed by a permutation-invariant aggregation. Set Transformer \cite{lee2019set} further improves model expressiveness by replacing the aggregation with self-attention to capture dependencies between set elements.
Meanwhile, many works focus on output-set modeling, especially in natural language processing, where the input is a sequence and the output is an unordered set of labels. A key challenge here is the mismatch between sequential decoding and unordered set targets. Some methods address this through order-invariant objectives or set-level rewards \cite{yang2019deep, cao2022otseq2set}, while others marginalize over label permutations or incorporate set cardinality information \cite{qin2019adapting, madaan2022conditional, tang2022sequence}.

% Overall, existing set modeling studies can be broadly divided into two categories: input-set modeling, which aims to learn permutation-invariant representations from unordered inputs, and output-set modeling, which aims to resolve the mismatch between sequential generation and unordered sets.

% -------------notations table-----
\begin{table}
\centering
\caption{Key notations.}
\label{tab:notations}
\small
\setlength{\tabcolsep}{4pt}
\renewcommand{\arraystretch}{1.15}
\begin{tabular}{l l}
\toprule
Notation & Meaning \\
\hline
$\mathcal{G}$     &     The knowledge graph  \\
$\mathcal{E}$     &   All entities in the knowledge graph \\
$\mathcal{R}$     &   All relations in the knowledge graph \\
$\mathcal{T}$     &   All triplets in the knowledge graph \\
$e,r$             &   Entity, relation \\
$\mathcal{R}^+(e)$            &   The set of compatible relations of $e$ \\
$ \mathcal{R}^-(e)$          &   The set of incompatible relations of $e$ \\
$\mathcal{R}_o(e)$            &   The set of observed relations of $e$ \\
$\mathcal{R}_m(e)$           &   The set of missing relations of $e$ \\
$\mathcal{R}_c(e)$           &   The set of candidate relations of $e$ \\
$\mathcal{N}(e)$ &   Negative sampled relations of $e$ \\
$s(r|\mathcal{R}_o(e))$       &   The score of $r$ given ${R}_o(e)$ \\
\bottomrule
\end{tabular}
\end{table}
% ---------------------------------

\section{Problem Formulation}
\label{sec:Prob_def}

In this section, we introduce the key notation used throughout the paper, and provide the definition of the proposed RSC problem.
A knowledge graph (KG) is denoted as $\mathcal{G} = (\mathcal{E}, \mathcal{R}, \mathcal{T})$, where $\mathcal{E}$, $\mathcal{R}$, and $\mathcal{T}$ denote the sets of entities, relations, and triplets, respectively. Each triplet $(h, r, t) \in \mathcal{T}$ includes a head entity $h \in \mathcal{E}$, a tail entity $t \in \mathcal{E}$, and the relation that connects them $r \in \mathcal{R}$. 
We embed a relation $r$ into the vector space and denote it as $\boldsymbol{r} \in \mathbb{R}^d$, where $d$ is the embedding dimension.
Given an entity $e \in \mathcal{E}$, all the relations in $\mathcal{R}$ can be divided into two sets, the compatible relation set $\mathcal{R}^+(e)$ and the incompatible relation set $\mathcal{R}^-(e)$, and there is:
\begin{equation}
\begin{aligned}
\mathcal{R} &= \mathcal{R}^+(e) \cup \mathcal{R}^-(e)\\ 
\emptyset   &= \mathcal{R}^+(e) \cap \mathcal{R}^-(e).
\end{aligned}
\end{equation}
We denote the observed relation set of entity $e$ as $\mathcal{R}_o(e)$, and it includes relations that have been associated with it through observed triplets in the KG:
\begin{equation}
\mathcal{R}_o(e) = \{ r \in \mathcal{R} \mid \exists (h, r, t) \in \mathcal{T}, \; h = e \lor t = e \},
\end{equation}
Accordingly, we denote the missing relation set of entity $e$ as $\mathcal{R}_m(e)$. It includes relations that are compatible with $e$ but have not been associated with $e$ in the KG, and there is:
\begin{equation}
\mathcal{R}^+(e) = \mathcal{R}_m(e) \cup \mathcal{R}_o(e).
\end{equation}
Then, we define the RSC problem as: 
\begin{equation}
f: (\mathcal{G},e)\rightarrow \mathcal{R}_m(e),
\end{equation}
where $f$ is the model to be learned. Given a target entity $e$ and the KG, the goal of RSC is to learn $f$ such that the predicted missing relation set approximates the true missing relation set $\mathcal{R}_m(e)$ as closely as possible.

\section{Methodology}
In this section, we present the proposed RelSetE method. RelSetE formulates RSC as predicting missing compatible relations conditioned on the observed relations of a target entity, based on the premise that compatible relations of an entity exhibit semantic dependencies that can be exploited to infer missing ones.

\subsection{Overview}
\label{subsec:overview}
RelSetE resolves the RSC problem by adopting the semantic dependencies among relations. For an entity $e$, it measures the compatibility score of a candidate relation $r$ as follows:
\begin{equation}
s(r \mid \mathcal{R}_o(e)) : \mathcal{E} \times \mathcal{R} \rightarrow \mathbb{R}
\end{equation}
Candidates with the highest scores will be regarded as predicted missing relations.
RelSetE consists of three components: a relation embedding layer, a permutation-invariant encoder, and a compatibility scoring decoder. Given an entity $e$ with observed relation set $\mathcal{R}_o(e)$, the model proceeds as follows.
First, each relation is mapped to a dense vector via the relation embedding layer. The encoder takes the relation embedding set of $\mathcal{R}_o(e)$ as input and aggregates them into a single relation-aware hidden representation $\boldsymbol{z} \in \mathbb{R}^d$. Specifically, to preserve the permutation-invariant nature of relation sets, the encoder first obtains contextualized relation embeddings with multi-head attention, which capture interactions among observed relations, and then adopts attention pooling to aggregate contextualized relation embeddings into the hidden representation $\boldsymbol{z}$ for $\mathcal{R}_o(e)$ in a permutation-invariant manner. Finally, the decoder estimates the compatibility score of each candidate relation by measuring its similarity with $\boldsymbol{z}$, where a higher score indicates greater semantic compatibility with the entity. The overview of RelSetE is illustrated in Figure~\ref{fig:RelSetE}.

\subsection{Encoder}
\label{subsec:encoder}
For an entity $e$, the encoder of RelSetE encodes its observed relation set $\mathcal{R}_o(e)$ into the hidden representation $\boldsymbol{z}$. We denote the relation embedding set of $\mathcal{R}_o(e)$ as $\boldsymbol{R} = \{\boldsymbol{r}_1;...;\boldsymbol{r}_{|\mathcal{R}_o(e)|}\}$, where $r_{i} \in \mathcal{R}_o(e)$.
As discussed, a set is inherently permutation-invariant, i.e., the information it encodes is independent of the order in which its elements are arranged.
The attention mechanism is naturally suited to this requirement, as each output is computed as a weighted sum over all input elements where the weights depend only on pairwise key-query similarities. As a result, any permutation of the inputs permutes the keys and values consistently, leaving the output unchanged. This makes attention a principled choice for aggregating relational information in a permutation-invariant manner.
Specifically, we adopt scaled dot-product attention \cite{vaswani2017attention} as follows:

\begin{equation}
    \mathrm{Att}(Q, K, V)  = \mathrm{softmax}(\frac{QK^T}{\sqrt{d_K}}) V,
\end{equation}
where $Q$, $K$, $V$ are query, key and value, $d_K$ is the dimension of the key vectors. 

Based on this, we adopt multi-head attention and attention pooling to obtain the permutation-invariant hidden representation $\boldsymbol{z}$.
Specifically, the attention of $\mathcal{R}_o(e)$ is computed as follows:
\begin{equation}
\begin{aligned}
    head = \mathrm{Att}\!\left(
    \boldsymbol{R} W_i^Q,\;
    \boldsymbol{R} W_i^K,\;
    \boldsymbol{R} W_i^V\
    \right),
    \label{eq:each_head_attention}
\end{aligned}
\end{equation}
where $W^Q, W^K, W^V \in \mathbb{R}^{d \times m \times \frac{d}{m}}$ are trainable projection matrices. The contextualized relation representation is then obtained by:
\begin{equation}
        H = \mathrm{concat}(head_1, \ldots, head_m) \cdot W^l,
\end{equation}
where $W^l$ is a linear transformation, and $H \in \mathbb{R}^{|\mathcal{R}_o(e)|\times d}$ represents contextualized relation representations for $\mathcal{R}_o(e)$.

Attention pooling is then applied to $H$ to generate $\boldsymbol{z}$ as follows: 
\begin{equation}
    \boldsymbol{z} = \mathrm{Att}(\boldsymbol{\omega}, H, H)  = \mathrm{softmax}(\frac{\boldsymbol{\omega} H^T}{\sqrt{d_H}}) H,
\end{equation}
where $\boldsymbol{\omega} \in \mathbb{R}^{d}$ is a trainable seed vector. 
In this way, RelSetE encodes the information in 
$\mathcal{R}_o(e)$ into $\boldsymbol{z}$.

\subsection{Decoder}
\label{subsec:decoder}
% The decoder's behaviors is slightly different during optimization and inference
With the observed relation set $\mathcal{R}_o(e)$ of entity $e$, the decoder predicts missing relations by measuring the compatibility of candidate relations in $\mathcal{R}_c(e)$, which is defined as:
\begin{equation}
    \mathcal{R}_c(e) = \mathcal{R}\setminus \mathcal{R}_o(e).
\end{equation}
That is, the decoder queries $\boldsymbol{z}$ (yielded from the encoder) to measure the compatibility score of a candidate relation $r$ with the dot product as follows:
\begin{equation}
\label{eq:scores_pos_neg}
    s(r \mid \mathcal{R}_o(e)) = \boldsymbol{z} \cdot \boldsymbol{r}, \textbf{ }r \in \mathcal{R}_c(e).
\end{equation}
A higher compatibility score means the candidate relation is more likely to be compatible with the entity. We adopt such a measurement for all candidate relations $r\in \mathcal{R}_c(e)$ and obtain their compatibility scores. 
Candidate relations with top-k ranked compatibility scores are regarded as the predicted missing relations.

\subsection{Learning Objective}
\label{subsec:learning_obj}
As explained above, we expect the model to produce higher scores for compatible relations and lower scores for incompatible relations.
That is, for any $r \in \mathcal{R}^+(e)$ and $r^- \in \mathcal{R}^-(e)$, we aim to satisfy:
\begin{equation}
s(r \mid \mathcal{R}_o(e)) > s(r^- \mid \mathcal{R}_o(e)).
\end{equation}
Training the model relies on the ground-truth set $\mathcal{R}_m(e)$.
However, they are challenging to obtain in real-world scenarios. Therefore, we refer to the paradigm of self-supervised learning for effective model training. During training, for an entity $e$, its observed relations $\mathcal{R}_o(e)$ are randomly partitioned into a pseudo-observed set $\mathcal{C}(e)$ and a pseudo-missing compatible relation set $\mathcal{P}(e)$. Then, the objective is to maximize the score of $s(r \mid \mathcal{C}(e)),r\in \mathcal{P}(e)$. In addition, we introduce negative samples into the objective to further improve the discriminative ability for compatible relations and incompatible relations. The negative samples are randomly sampled from $\mathcal{R}_c(e)$, i.e., $\mathcal{N}(e)=\{r^{-}|r^{-}\sim \mathcal{R}_c(e)\}$, where the probability of sampling a missing relation is negligible given the size of the candidate set. The learning objective is:
\begin{equation}
\label{eq:loss}
\begin{aligned}
\mathcal{L}(\theta)
&=
\sum_{e \in \mathcal{E}} \
\sum_{\mathcal{P}(e) \subseteq \mathcal{R}_o(e)}
\frac{1}{|\mathcal{P}(e)|}
\sum_{r \in \mathcal{P}(e)}
\\
&\quad
\left[
-\log
\frac{
e^{s(r \mid \mathcal{C}(e))/\tau}
}{
e^{s(r \mid \mathcal{C}(e))/\tau}
+
\sum\limits_{r^- \in \mathcal{N}(e)}
e^{s(r^- \mid \mathcal{C}(e))/\tau}
}
\right],
\end{aligned}
\end{equation}
where $\tau > 0$ is a temperature parameter. We adopt multiple random partitions for $\mathcal{R}_o(e)$ to obtain different $\mathcal{C}(e)$ and $\mathcal{P}(e)$ in each training round to enhance model generalization. The entire RelSetE model is trained end-to-end via standard backpropagation.

\begin{figure*}[h]
    \centering
    \begin{subfigure}[t]{0.32\linewidth}
        \centering
        \includegraphics[width=\linewidth]{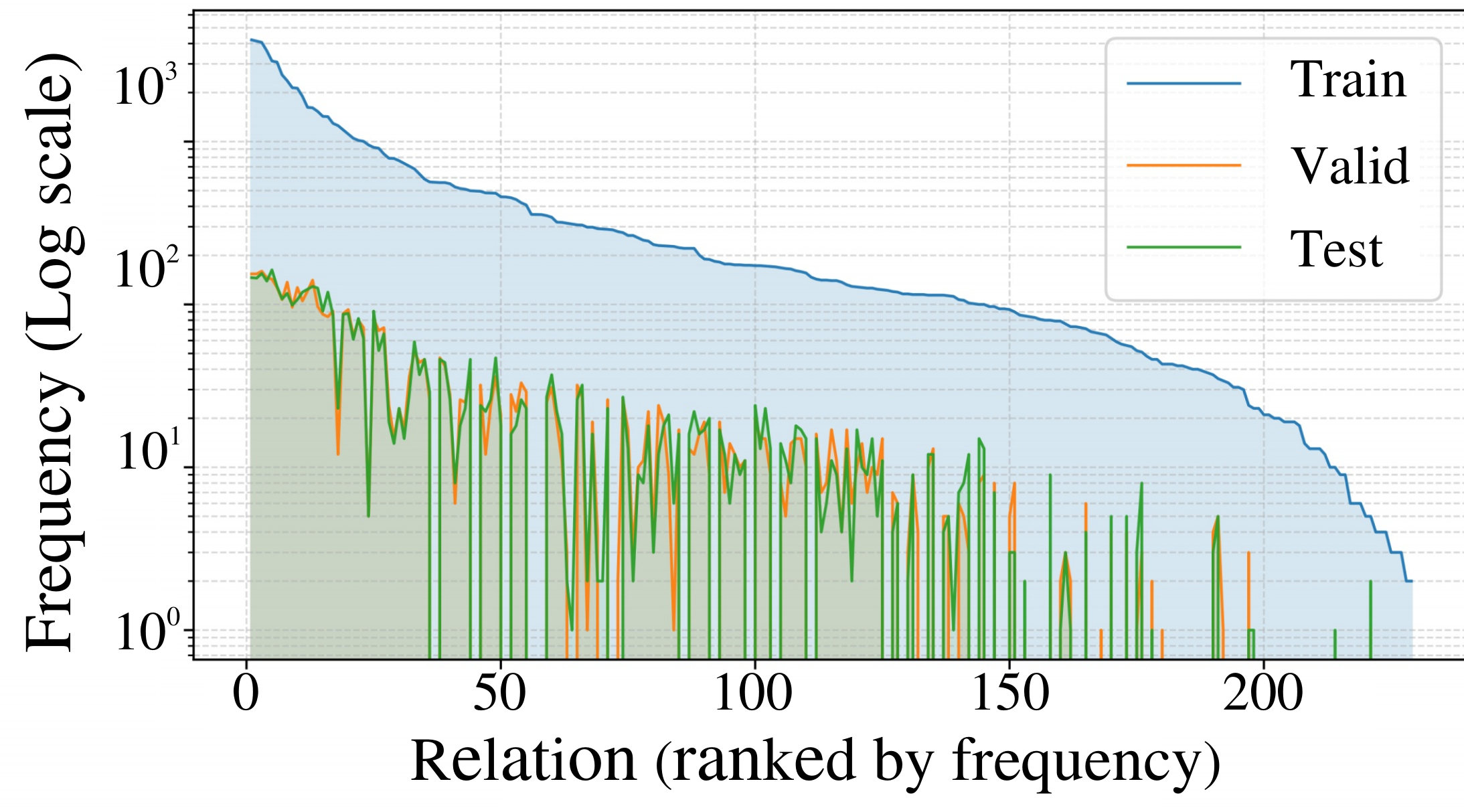}
        \caption{FB15k-237-re}
        \label{fig:fb_rel_dist}
    \end{subfigure}\hfill
    \begin{subfigure}[t]{0.29\textwidth}
        \centering
        \includegraphics[width=\linewidth]{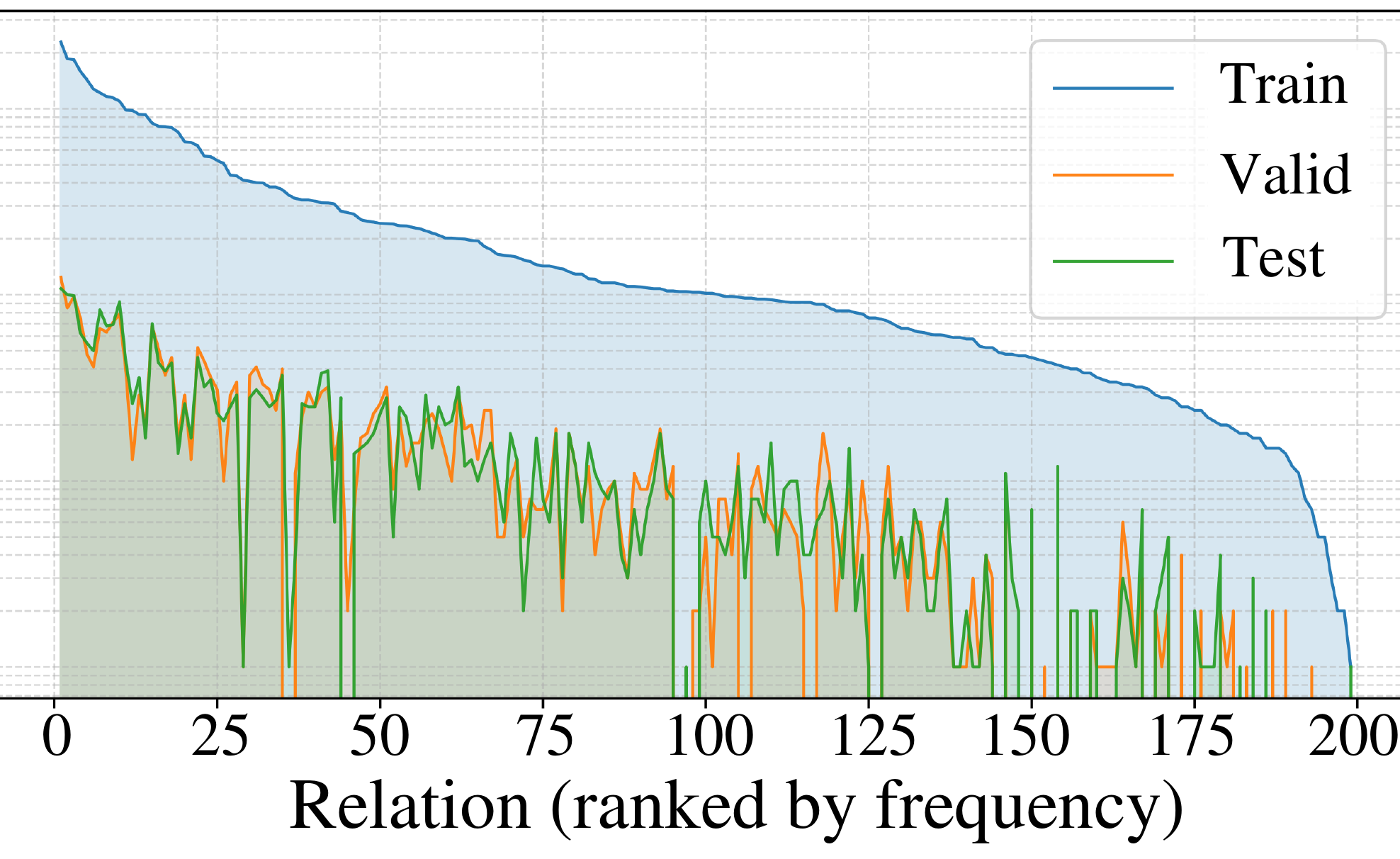}
        \caption{NELL-995-re}
        \label{fig:nell995_real_dist}
    \end{subfigure}\hfill
    \begin{subfigure}[t]{0.2915\textwidth}
        \centering
        \includegraphics[width=\linewidth]{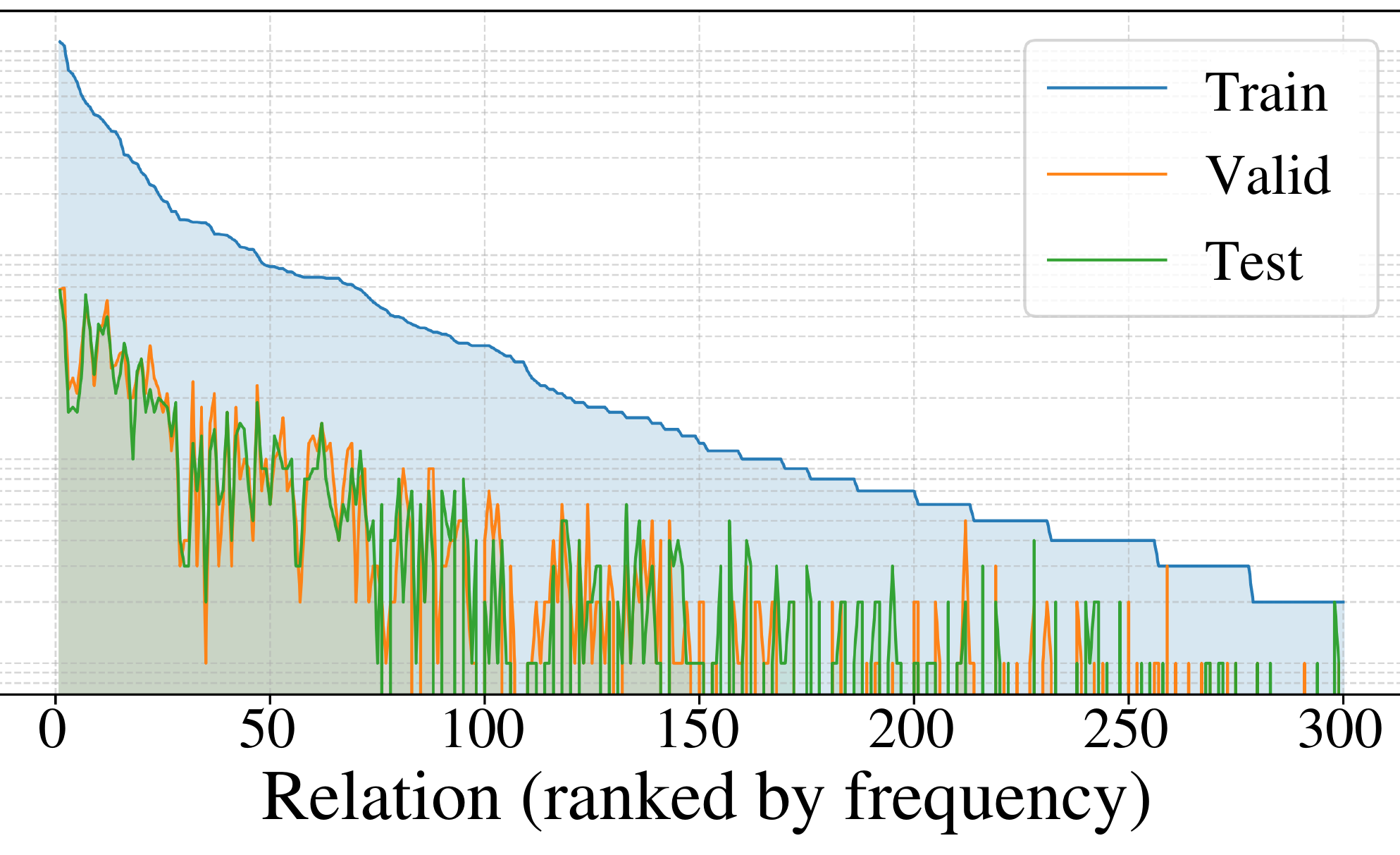}
        \caption{NELL-1115-re}
        \label{fig:nell1115_rel_dist}
    \end{subfigure}
    \caption{Relation frequency distribution of each reconstructed dataset.}
    \label{fig:dataset_stats}
\end{figure*}

\section{Experiments and Results}
\label{sec:experiments}
\subsection{Dataset}
To evaluate the proposed RelSetE method for the RSC problem, we reconstruct three datasets from existing KGC datasets, and they are FB15k-237-re, NELL-995-re, and NELL-1115-re. Specifically, FB15k-237-re and NELL-995-re are derived from FB15k-237~\cite{toutanova2015observed} and NELL-995~\cite{nell995}, which are KG benchmarks extracted from Freebase, which covers entities such as people, places, films, and organizations, and the NELL system, which is automatically constructed from web text spanning multiple domains, respectively. NELL-1115-re is derived from the large-scale NELL-08m-1115 dataset \cite{carlson2010toward} by selecting information-rich and high-degree subgraphs. 
The dataset reconstruction is processed by the following steps. We first derive the relation set of each entity from its associated triplets in the KG, collecting all relations that appear in triplets involving that entity. Entities with fewer than 2 relations are removed since they provide insufficient relation context for RSC. 
For the remaining entities, their relation sets are partitioned into training, validation, and test splits at the dataset level, maintaining an approximate 8:1:1 ratio.

\begin{table}[htbp]
\centering
\footnotesize

\caption{Statistics of the three reconstructed datasets, including dataset size, the relation set size for each entity ($|\mathcal{R}(e)|$), and the number of entities associated with each relation ($|\mathcal{E}(r)|$).}
\setlength{\tabcolsep}{4pt}
\renewcommand{\arraystretch}{1.15}

\begin{tabular}{llcc|cc|cc}
\toprule
\multicolumn{2}{c}{\multirow{2}{*}{Dataset}}& \multicolumn{2}{c}{Size}
 & \multicolumn{2}{c}{$|\mathcal{R}(e)|$}
 & \multicolumn{2}{c}{$|\mathcal{E}(r)|$} \\
\cmidrule(lr){3-4} \cmidrule(lr){5-6} \cmidrule(lr){7-8}
& &$\mid\mathcal{E}\mid$& $\mid\mathcal{R}\mid$ & Range & Mean & Range & Mean \\
\midrule

\multirow{3}{*}{\makecell{FB15k-\\237-re}}
                            & Train & 11,067 & 229 & 3$\sim$24 & 8.03 & 2$\sim$4,212 & 388.22 \\
                            & Valid & 1,928  & 146 & 2$\sim$ 5 & 2.28 & 1$\sim$160 & 30.05  \\
                            & Test  & 1,927  & 146 & 2$\sim$ 5 & 2.29 & 1$\sim$163 & 30.16  \\
\midrule
\multirow{3}{*}{\makecell{NELL-\\995-re}}
                          & Train & 11,224 & 199 & 3$\sim$32 & 4.23 & 1$\sim$2,280 & 238.42 \\
                          & Valid & 1,164  & 164 & 2$\sim$8 & 2.33  & 1$\sim$124 & 16.54  \\
                          & Test  & 1,163  & 166 & 2$\sim$8 & 2.35  & 1$\sim$108 & 16.47  \\
\midrule
\multirow{3}{*}{\makecell{NELL-\\1115-re}}
                           & Train & 4,501  & 406 & 3$\sim$53 & 4.41 & 1$\sim$1,110 & 48.89 \\
                           & Valid & 762   & 240 & 1$\sim$14 & 2.02 & 1$\sim$67 & 6.42   \\
                           & Test  & 761   & 222 & 1$\sim$11 & 2.15 & 1$\sim$69 & 7.36   \\
\bottomrule
\label{tab:dataset_stats}

\end{tabular}
\end{table}

The statistics of the three reconstructed datasets are summarized in Table~\ref{tab:dataset_stats}. 
Generally, the training sets of FB15k-237-re, NELL-995-re, and NELL-1115-re contain rich relation types, providing sufficient relation diversity for the evaluation. Among the three datasets, FB15k-237-re exhibits the richest relation sets per entity ($|\mathcal{R}(e)|$), while NELL-995-re and NELL-1115-re have comparatively fewer relations per entity. All three datasets exhibit a long-tail relation distribution, where a significant proportion of relations are associated with only a small number of entities, posing additional challenges for RSC. 
The detailed relation frequency distributions of the three datasets are shown in Figure~\ref{fig:dataset_stats}, revealing a highly imbalanced pattern where a small number of relations account for the majority of entity associations, while most relations appear only rarely. This imbalance makes it particularly challenging to predict infrequent relations, as the model has limited exposure to them during training. 

% --------------------------------
\begin{table*}[t]
\centering
\caption{The performance of compared models on FB15k-237-re, NELL-995-re, and NELL-1115-re. Each cell reports the mean with the variance underneath. Best results are in bold.}
\label{tab:results_all}
\small
\renewcommand{\arraystretch}{1.15}
\setlength{\tabcolsep}{2.6pt}

\newlength{\metricwidth}
\setlength{\metricwidth}{1.45cm}

\begin{tabular}{l >{\centering\arraybackslash}m{1.5cm}>{\centering\arraybackslash}m{1.5cm}>{\centering\arraybackslash}m{1.5cm}|
>{\centering\arraybackslash}m{1.5cm}>{\centering\arraybackslash}m{1.5cm}>{\centering\arraybackslash}m{1.5cm}|
>{\centering\arraybackslash}m{1.5cm}>{\centering\arraybackslash}m{1.5cm}>{\centering\arraybackslash}m{1.5cm}}
\toprule
\multirow{2}{*}{\textbf{Model}}
& \multicolumn{3}{c}{\textbf{FB15k-237-re}}
& \multicolumn{3}{c}{\textbf{NELL-995-re}}
& \multicolumn{3}{c}{\textbf{NELL-1115-re}} \\
& Precision & Recall & F1-score
& Precision & Recall & F1-score
& Precision & Recall & F1-score \\

\midrule
% \multirow{2}[c]{*}{}
\multirow[c]{2}{*}{RNN~\cite{vinyals2015order}}
&  0.3198
&  0.1453
&  0.1984
&  0.1422
&  0.0627
&  0.0856
&  0.0639
&  0.0440
&  0.0500 \\
&$^{\pm1.7{\times}10^{-3}}$
&$^{\pm3.3{\times}10^{-4}}$
&$^{\pm6.3{\times}10^{-4}}$
&$^{\pm1.6{\times}10^{-3}}$
&$^{\pm3.8{\times}10^{-4}}$
&$^{\pm6.7{\times}10^{-4}}$
&$^{\pm3.9{\times}10^{-4}}$
&$^{\pm2.8{\times}10^{-4}}$
&$^{\pm3.0{\times}10^{-4}}$ \\
\multirow[c]{2}{*}{RPW RNN~\cite{vinyals2015order}}
& 0.4965
& 0.4860
& 0.4265
& \textbf{0.4865}
& 0.2259
& 0.3042

& 0.4142 	
& 0.1871 	
& 0.2544 		
\\
&$^{\pm5.3{\times}10^{-2}}$
&$^{\pm5.4{\times}10^{-2}}$
&$^{\pm1.0{\times}10^{-5}}$
&$^{\pm4.4{\times}10^{-5}}$
&$^{\pm3.4{\times}10^{-5}}$
&$^{\pm1.4{\times}10^{-5}}$
&$^{\pm1.4{\times}10^{-2}}$
&$^{\pm4.2{\times}10^{-3}}$
&$^{\pm7.2{\times}10^{-3}}$ \\

\multirow[c]{2}{*}{Seq2Set~\cite{yang2019deep}}
& 0.1774
& 0.2712
& 0.2020
& 0.0702
& 0.1801
& 0.0936
& 0.0197
& 0.0726
& 0.0296 \\
&$^{\pm1.9{\times}10^{-3}}$
&$^{\pm1.7{\times}10^{-3}}$
&$^{\pm3.4{\times}10^{-4}}$
&$^{\pm7.0{\times}10^{-4}}$
&$^{\pm9.5{\times}10^{-4}}$
&$^{\pm4.5{\times}10^{-4}}$
&$^{\pm6.4{\times}10^{-5}}$
&$^{\pm7.3{\times}10^{-4}}$
&$^{\pm1.4{\times}10^{-4}}$ \\

\multirow[c]{2}{*}{OTSeq2Set~\cite{cao2022otseq2set}}
& 0.3928
& 0.1779
& 0.2331
& 0.1493
& 0.0650
& 0.0892
& 0.0792
& 0.0534
& 0.0670 \\
&$^{\pm7.0{\times}10^{-3}}$
&$^{\pm1.4{\times}10^{-3}}$
&$^{\pm2.6{\times}10^{-3}}$
&$^{\pm2.7{\times}10^{-4}}$
&$^{\pm7.0{\times}10^{-5}}$
&$^{\pm1.2{\times}10^{-4}}$
&$^{\pm2.7{\times}10^{-4}}$
&$^{\pm6.7{\times}10^{-5}}$
&$^{\pm4.1{\times}10^{-5}}$ \\

\multirow[c]{2}{*}{DeepSet~\cite{zaheer2017deep}}
& 0.3729
& 0.4784
& 0.4159
& 0.1589
& 0.2798
& 0.1995
& 0.0765
& 0.1709
& 0.0995 \\
&$^{\pm1.2{\times}10^{-5}}$
&$^{\pm5.0{\times}10^{-4}}$
&$^{\pm2.0{\times}10^{-5}}$
&$^{\pm4.2{\times}10^{-6}}$
&$^{\pm1.6{\times}10^{-5}}$
&$^{\pm6.8{\times}10^{-6}}$
&$^{\pm1.2{\times}10^{-3}}$
&$^{\pm6.1{\times}10^{-3}}$
&$^{\pm2.1{\times}10^{-3}}$ \\

\multirow[c]{2}{*}{Set Transformer~\cite{lee2019set}}
& 0.3382 
& 0.4564 
& 0.3850 
& 0.2405
& 0.4178
& 0.3004
& 0.1883
& 0.4386
& 0.2506 \\
& $^{\pm3.5{\times}10^{-4}}$ 
& $^{\pm7.0{\times}10^{-4}}$ 
& $^{\pm4.9{\times}10^{-4}}$
&$^{\pm5.5{\times}10^{-7}}$
&$^{\pm3.3{\times}10^{-10}}$
&$^{\pm8.1{\times}10^{-9}}$
&$^{\pm3.1{\times}10^{-4}}$
&$^{\pm3.8{\times}10^{-3}}$
&$^{\pm5.2{\times}10^{-3}}$ \\

\multirow[c]{2}{*}{MLC~\cite{zhao2022well}}
& 0.3735
& 0.5047
& 0.4254
& 0.3344 
& 0.4408
& 0.3745
& 0.1257
& 0.2842
& 0.1650 \\
&$^{\pm4.6{\times}10^{-3}}$
&$^{\pm8.3{\times}10^{-4}}$
&$^{\pm4.7{\times}10^{-3}}$
&$^{\pm3.8{\times}10^{-5}}$
&$^{\pm8.7{\times}10^{-5}}$
&$^{\pm5.7{\times}10^{-5}}$
&$^{\pm5.5{\times}10^{-3}}$
&$^{\pm2.8{\times}10^{-2}}$
&$^{\pm9.7{\times}10^{-3}}$ \\

\midrule
\multirow[c]{2}{*}{\textbf{RelSetE}}
& \textbf{0.5993} 
& \textbf{0.5391} 
& \textbf{0.5625}
& 0.3755 
& \textbf{0.4905} 
& \textbf{0.4185}
& \textbf{0.4395}
& \textbf{0.5159}
& \textbf{0.4496} \\

&$^{\pm9.3{\times}10^{-4}}$ 
&$^{\pm1.5{\times}10^{-3}}$ 
&$^{\pm1.3{\times}10^{-3}}$ 
&$^{\pm8.1{\times}10^{-3}}$ 
&$^{\pm1.2{\times}10^{-2}}$ 
&$^{\pm9.6{\times}10^{-3}}$
&$^{\pm2.9{\times}10^{-4}}$
&$^{\pm3.3{\times}10^{-4}}$
&$^{\pm2.8{\times}10^{-4}}$ \\
\bottomrule
\end{tabular}
\end{table*}

% --------------------------------------------------

\subsection{Baseline Models}
\label{subsec:baseline}
To evaluate RelSetE, we compare it against baseline methods spanning three categories of set modeling paradigms: sequence-to-sequence, sequence-to-set, and set-to-set. For sequence-to-sequence, we include RNN and RPW RNN~\cite{vinyals2015order}; for sequence-to-set, we include Seq2Set~\cite{yang2019deep} and OTSeq2Set~\cite{cao2022otseq2set}; for set-to-set, we include DeepSet~\cite{zaheer2017deep}, Set Transformer~\cite{lee2019set}, and multi-label classification (MLC)~\cite{zhao2022well}.
\begin{itemize}
    \item \textbf{RNN and RPW RNN}~\cite{vinyals2015order} adopt vanilla RNN for the input and generate output in an auto-regressive manner. RPW RNN further introduces set-based memory processing to reduce sensitivity to input orders.
    \item \textbf{Seq2Set}~\cite{yang2019deep} adopts sequential encoding and addresses the mismatch between sequential decoding and unordered targets through permutation-invariant loss optimization.
    \item \textbf{OTSeq2Set}~\cite{cao2022otseq2set} extends Seq2Set by incorporating optimal transport theory to better align sequential predictions with unordered target sets, enabling more robust matching between generated outputs and ground-truth labels.
    \item \textbf{DeepSet}~\cite{zaheer2017deep} proposes a permutation-invariant neural network framework for set-structured inputs, processing each element independently and aggregating representations via a symmetric pooling function to ensure invariance to input ordering.
    \item \textbf{Set Transformer}~\cite{lee2019set} introduces an attention-based framework for permutation-invariant processing of set-structured inputs, leveraging self-attention and induced set attention blocks to efficiently capture interactions among set elements.
    \item \textbf{MLC}~\cite{zhao2022well} proposes a modified loss function that assigns higher weights to confidently correct predictions to improve classification performance. We set its encoder to the same as RelSetE for fair comparison.
\end{itemize}

\subsection{Experimental Setup}
\label{subsec:experiments_settings}
For all experiments, we use a batch size of 128 and train all models using the Adam optimizer with a learning rate of $1\times 10^{-3}$ and gradient clipping at a maximum norm of 1.0. For DeepSet, Set Transformer, and RelSetE, the number of negative relations is fixed as 32. Unless otherwise specified, the dropout rate, relation embedding dimension, and hidden dimension are set to 0.2, 256, and 256, respectively.
The hyperparameters for each baseline are configured as follows. For RNN and RPW RNN, we use 256-dimensional embeddings, a hidden dimension of 512, 2 recurrent layers, a dropout rate of 0.1, and a teacher forcing ratio of 0.2; RPW RNN additionally replaces the recurrent encoder with an RPW module. Seq2Set and OTSeq2Set share the same recurrent backbone with identical hyperparameters, but differ in their training objectives: Seq2Set optimizes a reinforcement learning objective based on set-level reward, while OTSeq2Set employs bipartite matching followed by cross-entropy loss. For DeepSet, Set Transformer, and RelSetE, the same scoring decoder is shared across all three models. DeepSet uses 256-dimensional relation embeddings with mean pooling; Set Transformer employs 2 set-attention layers with 2 attention heads; RelSetE uses 2 multi-head attention layers with 2 attention heads and a trainable attention-pooling seed vector of dimension 256. For MLC, a 256-dimensional relation-aware representation is passed to a classification head with hidden layers of dimensions 512 and 256, trained with binary cross-entropy (BCE) loss.

% -------------
\subsection{Main Results}
\label{subsec:main_results}

Table~\ref{tab:results_all} reports RSC performance of the proposed RelSetE and benchmark models on FB15k-237-re, NELL-995-re, and NELL-1115-re. Overall, RelSetE achieves the best F1-score on all three datasets, demonstrating consistent effectiveness for the relation set completion.

On FB15k-237-re, NELL-995-re, and NELL-1115-re, RelSetE achieves F1-scores of 0.5625, 0.4185, and 0.4496, respectively, consistently outperforming all baselines across all three datasets. The margins are particularly pronounced on FB15k-237-re and NELL-1115-re, where the strongest baselines, RPW RNN and Set Transformer, reach only 0.4265 and 0.2544, respectively. These results demonstrate that RelSetE delivers robust prediction quality across datasets with varying relation distributions.
Beyond F1, RelSetE also maintains a favorable precision-recall balance. On FB15k-237-re and NELL-1115-re, RelSetE leads in all three metrics simultaneously, indicating improvements in both prediction correctness and ground-truth coverage. On NELL-995-re, RPW RNN attains higher precision than RelSetE, but its substantially lower recall yields a weaker F1-score, suggesting a tendency to predict fewer but more conservative relations.

The comparison with sequence-to-sequence models (RNN, RPW RNN) highlights the limitation of imposing artificial orders on relation sets. The RNN baseline performs weakly across all datasets, especially on NELL-995-re and NELL-1115-re, where its F1-scores are 0.0856 and 0.0500, respectively. RPW RNN improves over RNN by reducing input-order sensitivity, but it still relies on auto-regressive decoding and therefore retains an enforced output-order. Although RPW RNN performs competitively on FB15k-237-re and NELL-995-re, its recall remains substantially lower than that of RelSetE, and its performance drops on NELL-1115-re. 
Sequence-to-set methods, including Seq2Set and OTSeq2Set, are designed to address unordered output sets by proposing an order-invariant loss function, while retaining the sequential input.
They do not perform competitively in this task. Their relatively low F1-scores indicate that output-order correction alone does not fully address the RSC problem, where the input itself is also a set of relations.
Set-based baselines confirm the benefit of permutation-invariant input modeling. DeepSet and Set Transformer generally achieve higher recall than sequence-based baselines, validating the importance of respecting the unordered nature of relation sets. However, their F1-scores remain below those of RelSetE across all datasets.

\begin{table}[htbp]
\centering
\caption{The performance of RelSetE regarding different relation frequencies ($|\mathcal{E}(r)|$).}
\label{tab:entity_relation_pair_count}
\footnotesize
\setlength{\tabcolsep}{7pt}
\renewcommand{\arraystretch}{1.05}
\begin{tabular}{lcccc}
\toprule
\textbf{Dataset} & $\boldsymbol{|\mathcal{E}(r)|}$ & \textbf{Precision} & \textbf{Recall} & \textbf{F1-score} \\
\midrule 

\multirow[c]{8}{*}{FB15k-237-re}
    & 5$\sim$97       & 0.2388 & 0.2313 & 0.2338 \\
    & 100$\sim$128    & 0.3057 & 0.4157 & 0.3380 \\
    & 129$\sim$175    & 0.3732 & 0.4645 & 0.4006 \\
    & 175$\sim$265    & 0.2878 & 0.4731 & 0.3467 \\ 
    & 276$\sim$482    & 0.5302 & 0.6398 & 0.5631 \\
    & 484$\sim$840    & 0.5722 & 0.8503 & 0.6556 \\
    & 910$\sim$2,567   & 0.4930 & 0.9701 & 0.6263 \\
    & 3,078$\sim$4,212  & 0.8117 & 0.9524 & 0.8573 \\
\addlinespace[2pt]
\midrule
\multirow[c]{8}{*}{NELL-995-re}
    & 1$\sim$40       & 0.3182 & 0.3409 & 0.3258  \\
    & 41$\sim$94      & 0.5953 & 0.6524 & 0.6087 \\
    & 95$\sim$113     & 0.6203 & 0.6466 & 0.6266 \\
    & 114$\sim$165    & 0.4700 & 0.6900 & 0.5353 \\    
    & 175$\sim$248    & 0.5657 & 0.7276 & 0.6135 \\   
    & 254$\sim$507    & 0.6565 & 0.7554 & 0.6824  \\
    & 508$\sim$1,219   & 0.4465 & 0.7362 & 0.5281 \\
    & 1,275$\sim$2,280  & 0.5146 & 0.7653 & 0.5818  \\
\addlinespace[2pt]
\midrule
\multirow[c]{8}{*}{NELL-1115-re}
    & 4$\sim$14        & 0.1966 & 0.1966 & 0.1966  \\
    & 15$\sim$21      & 0.3646 & 0.3854 & 0.3746 \\
    & 22$\sim$41      & 0.4712 & 0.4808 & 0.4742  \\
    & 41$\sim$72      & 0.5110 & 0.5623 & 0.5194  \\  
    & 72$\sim$89      & 0.5647 & 0.6744 & 0.5896  \\ 
    & 92$\sim$149     & 0.6206 & 0.7167 & 0.6504  \\
    & 164$\sim$488    & 0.4831 & 0.8905 & 0.5884  \\ 
    & 532$\sim$1,110   & 0.6620 & 0.9032 & 0.7377  \\
\bottomrule
\end{tabular}
\end{table}

\subsection{Effect of Relation Frequency}
We analyze the performance of RelSetE regarding relations with different frequency, i.e., $|\mathcal{E}(r)|$, which is the number of entities associated with relation $r$ in the dataset. For this, we rank all relations by their frequencies and evenly split them into 8 groups for the comparison, evaluating prediction accuracy within each group. The results are shown in Table~\ref{tab:entity_relation_pair_count}.

Generally, the results show higher prediction accuracy for relations of larger frequencies. 
In FB15k-237-re, recall increases substantially from 0.2313 for relations of the lowest frequency to 0.9524 for relations of the highest frequency.
The precision and F1-score show similar upward trends, rising from 0.2388 to 0.8117 and from 0.2338 to 0.8573, respectively.
A similar trend can also be observed in NELL-1115-re. That is, from relations of the lowest frequencies to relations of the highest frequencies, precision, recall, and F1-score increase from 0.1966 to 0.6620, 0.9032, and 0.7377, respectively.
For NELL-995-re, a similar improvement is observed, with precision, recall, and F1-score rising from 0.3182, 0.3409, and 0.3258 for the lowest-frequency relations to 0.5146, 0.7653, and 0.5818 for the highest-frequency relations, respectively.

These results are consistent with the intuition that relations with larger $|\mathcal{E}(r)|$ provide richer supervision during training, enabling RelSetE to better capture their semantic compatibility with entities and recover missing relations more effectively. Conversely, for less frequent relations, the limited training evidence makes it difficult for the model to learn reliable compatibility patterns, leading to degraded prediction performance. This performance gap highlights the inherent challenge posed by the long-tail relation distribution in RSC, and suggests that incorporating auxiliary structural or semantic information, such as entity neighborhood context or relation semantics from pre-trained language models, could be a promising direction to further improve performance on infrequent relations.

\begin{table}[h]
\centering
\caption{The performance of RelSetE under different entity relation-set sizes. ``No.'' denotes the number of entities with the corresponding relation-set size.}
\label{tab:degree_analysis}
\footnotesize
\setlength{\tabcolsep}{7pt}
\renewcommand{\arraystretch}{1.05}

\begin{tabular}{>{\centering\arraybackslash}m{1.4cm} >{\centering\arraybackslash}m{0.9cm} c c c c}
\toprule
\textbf{Dataset} & $\boldsymbol{|\mathcal{R}_o(e)|}$ & \textbf{No.} & \textbf{Precision} & \textbf{Recall} & \textbf{F1-score} \\
\midrule
\multirow[c]{5}{*}{FB15k-237-re}
        & 9$\sim$11  & 1210 & 0.4442 & 0.6661 & 0.5330 \\
        & 12$\sim$14 & 588  & 0.5199 & 0.5975 & 0.5509 \\
        & 15$\sim$17 & 109  & 0.5811 & 0.5344 & 0.5544 \\
        & 18$\sim$20 & 17   & 0.7059 & 0.5236 & 0.6008 \\
        & 22$\sim$24 & 3    & 0.6667 & 0.4000 & 0.5000 \\

\addlinespace[2pt]
\midrule
\multirow[c]{5}{*}{NELL-995-re}
          
        & 4$\sim$8   & 954 & 0.2452 & 0.7373 & 0.3671 \\
        & 9$\sim$13  & 158 & 0.3758 & 0.7221 & 0.4916 \\
        & 14$\sim$18 & 40  & 0.4750 & 0.6663 & 0.5523 \\
        & 19$\sim$24 & 8   & 0.6220 & 0.6512 & 0.6343 \\
        & 26$\sim$32 & 3   & 0.5000 & 0.4345 & 0.4635 \\
 
\addlinespace[2pt]
\midrule
\multirow[c]{5}{*}{NELL-1115-re}
        & 4$\sim$13 & 717 & 0.2220 & 0.7839 & 0.3365 \\
        & 14$\sim$23 & 29 & 0.4663 & 0.5845 & 0.5151 \\
        & 24$\sim$33 & 11 & 0.5303 & 0.4316 & 0.4741 \\
        & 34$\sim$43 & 2 & 0.6905 & 0.4500 & 0.5441 \\
        & 44$\sim$53 & 3 & 0.3333 & 0.1545 & 0.2109 \\
\bottomrule
\end{tabular}
\end{table}

%  ---------------------------------

\subsection{Effect of Relation-set Size}
\label{subsec:long_tail_analyse}
In this section, we analyze how RelSetE performs on entities with different relation-set sizes ($|\mathcal{R}_o(e)|$), where $|\mathcal{R}_o(e)|$ denotes the number of distinct types of relation compatible with entity $e$. For this, we evenly split entities into 6 groups by the relation-set size. As shown in Table~\ref{tab:degree_analysis}, the relation-set size of entities in the three datasets shows clear long-tail distributions, i.e., most entities have small relation sets due to the incompleteness of KGs.

Generally, the F1-score increases for entities of larger $|\mathcal{R}_o(e)|$. 
The F1-score increases from 0.5330 to 0.6008 on FB15k-237-re, from 0.3671 to 0.6343 on NELL-995-re, and from 0.3365 to 0.5441 on NELL-1115-re as $|\mathcal{R}_o(e)|$ grows. However, all three datasets exhibit a performance drop for entities with the largest $|\mathcal{R}_o(e)|$, which can be partly attributed to the small sample size in this group.
Meanwhile, Recall decreases as $|\mathcal{R}_o(e)|$ grows, likely due to the increasing complexity and rarity of relation patterns associated with larger relation sets. The highest Recall values, i.e., 0.6661, 0.7373, and 0.7839 on FB15k-237-re, NELL-995-re, and NELL-1115-re, respectively, all occur at the smallest $|\mathcal{R}_o(e)|$ group.
Notably, for the smallest $|\mathcal{R}_o(e)|$ groups, RelSetE still obtains meaningful prediction performance across all datasets. The F1-scores are 0.5330 for entities of $9\sim 11$ on FB15k-237-re, 0.3671 for entities of $4\sim 8$ on NELL-995-re, and 0.3365 for entities of $4\sim 13$ on NELL-1115-re. 
These groups contain the majority of entities in each dataset, with $62.7\%$, $82.0\%$, and $94.1\%$ entities, respectively. 
This shows that RelSetE is relatively effective at inferring missing compatible relations even for entities whose observed relation sets are relatively small.

\begin{figure}[h]
    \centering
    \includegraphics[width=\linewidth]{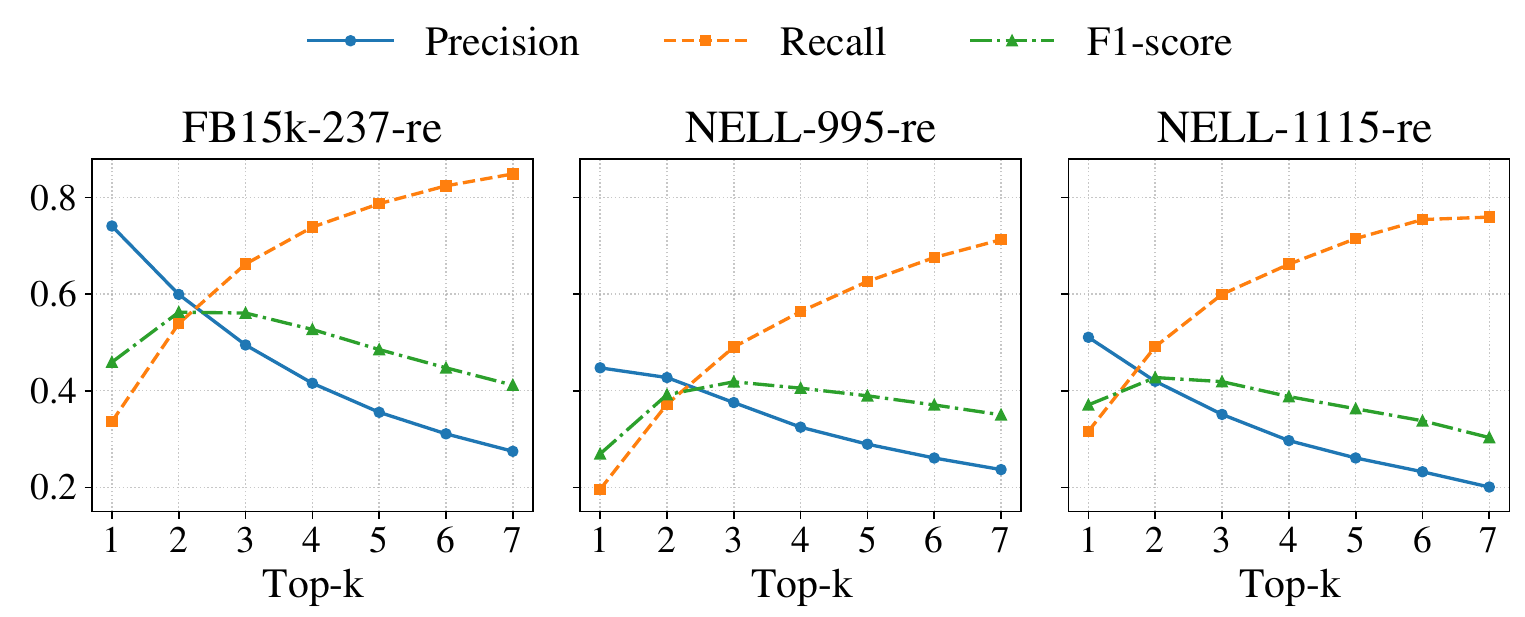}
    \caption{Top-k sensitivity of RelSetE.}
    \label{fig:topk_ana}
\end{figure}

\begin{figure}[h]
    \centering
    \includegraphics[width=\linewidth]{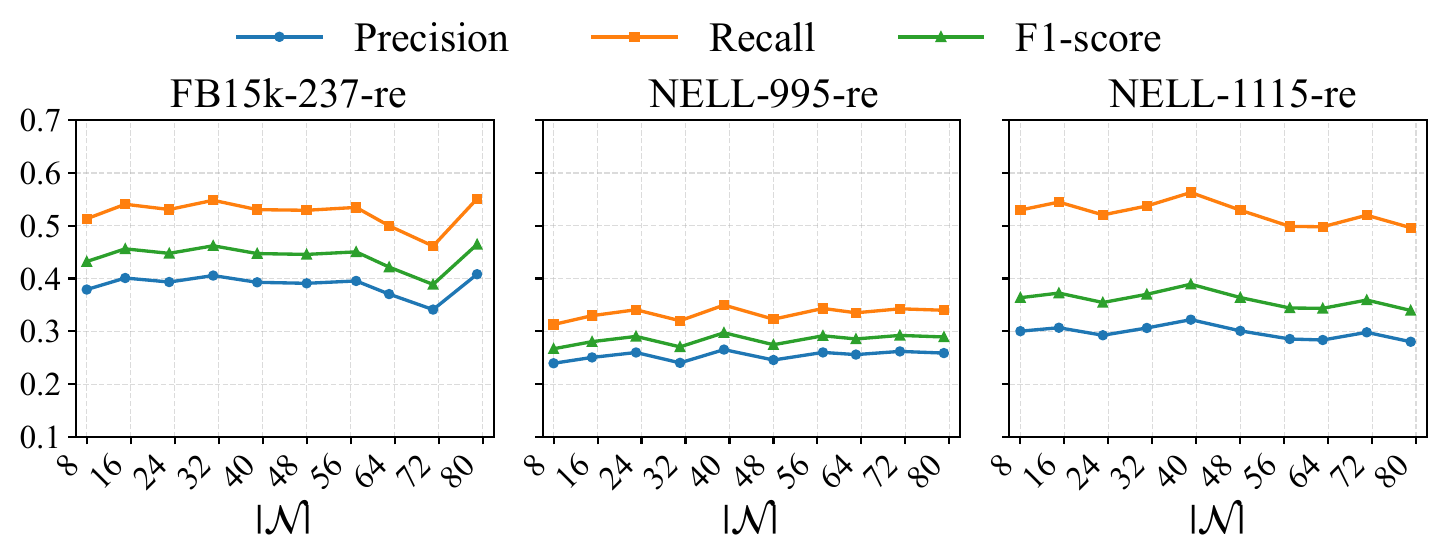}
    \caption{Negative sampling sensitivity of RelSetE, $|\mathcal{N}|$ is negative sampling size.}
    \label{fig:neg_sample_sensitivity}
\end{figure}

\subsection{Hyperparameter Analysis}
\label{subsec:arameter_analysis}
\textbf{Top-k sensitivity}:
Figure~\ref{fig:topk_ana} shows the sensitivity of RelSetE regarding top-k$\in \{1,\ldots,7\}$, which decides the number of predicted missing relations.
As k increases, Precision consistently decreases since more predicted relations introduce additional false positives, while Recall increases due to broader coverage of the ground-truth relation set. RelSetE achieves its best F1-score at a moderate k, specifically 2, 3, and 3 for FB15k-237-re, NELL-995-re, and NELL-1115-re, respectively, which aligns with the mean missing relation-set size reported in Table~\ref{tab:dataset_stats}. This suggests that RelSetE performs best when k matches the typical number of missing relations; beyond this point, additional predictions mostly introduce false positives, and the resulting loss in precision outweighs the gain in recall.

\textbf{Negative sampling sensitivity:}
Figure~\ref{fig:neg_sample_sensitivity} shows the performance of RelSetE regarding different negative sampling sizes $|\mathcal{N}|$ across FB15k-237-re, NELL-995-re, and NELL-1115-re. Overall, RelSetE shows stable performance under different negative sampling sizes. On FB15k-237-re, all three metrics remain stable with varying $|\mathcal{N}|$, with Precision around $0.34\sim0.41$, Recall around $0.46\sim0.55$, and F1-score around $0.39\sim0.47$. On NELL-995-re, F1-score stays around $0.27\sim0.30$, suggesting that the model is less sensitive to negative sampling on this dataset. On NELL-1115-re, the best F1-score appears around $|\mathcal{N}|=40$, after which the performance slightly decreases. 
These results indicate that a larger negative sampling size can provide richer contrastive supervision, but it may also introduce harder optimization or less informative negative comparisons. 

\subsection{Ablation Analysis}
\label{subsec:ablation_analysis}

% ------------------------tab Ablation --------------
\begin{table}[t]
\footnotesize
\centering
\caption{Ablation study on FB15k-237-re, NELL-995-re, and NELL-1115-re. Each cell reports the mean with the variance underneath.}
\label{tab:ablation_results_vertical}

\setlength{\tabcolsep}{1.8pt}
\renewcommand{\arraystretch}{1.2}
\setlength{\metricwidth}{1.5cm}

\newcommand{\mv}[2]{%
  \makebox[\metricwidth][c]{%
    \begin{tabular}[t]{@{}c@{}}
      #1\\[-1.10ex]
      \hspace{2.2em}{\scriptsize $\pm #2$}
    \end{tabular}%
  }%
}

\newcommand{\bmv}[2]{%
  \makebox[\metricwidth][c]{%
    \begin{tabular}[t]{@{}c@{}}
      \textbf{#1}\\[-1.10ex]
      \hspace{2.2em}{\scriptsize $\pm #2$}
    \end{tabular}%
  }%
}

\begin{tabular*}{\linewidth}{@{\extracolsep{\fill}}>{\centering\arraybackslash}m{1.0cm}
>{\centering\arraybackslash}m{1.12cm}
ccc@{}}
\toprule
Dataset & Model & Precision & Recall & F1-score\\
\midrule

\multirow[c]{8}{*}{FB15k-237-re}
& \multirow[c]{2}{*}{\textbf{RelSetE}}
& \textbf{0.4893}
& \textbf{0.6568}
& \textbf{0.5555} \\
&
& $^{1.0{\times}10^{-4}}$
& $^{2.0{\times}10^{-4}}$
& $^{1.0{\times}10^{-4}}$ \\

& \multirow[c]{2}{*}{w/o ENC}
& 0.3078
& 0.5482
& 0.3905 \\
&
& $^{6.0{\times}10^{-4}}$
& $^{2.0{\times}10^{-3}}$
& $^{1.0{\times}10^{-3}}$ \\

& \multirow[c]{2}{*}{w/o AP}
& 0.2994
& 0.5398
& 0.3818 \\
&
& $^{6.0{\times}10^{-7}}$
& $^{6.0{\times}10^{-7}}$
& $^{7.0{\times}10^{-7}}$ \\

& \multirow[c]{2}{*}{w/o MHA}
& 0.2831
& 0.5095
& 0.3608 \\
&
& $^{3.0{\times}10^{-5}}$
& $^{9.0{\times}10^{-5}}$
& $^{5.0{\times}10^{-5}}$ \\

% & \multirow[c]{2}{*}{MLC}
% & 0.3735
% & 0.5047
% & 0.4254 \\
% &
% &$^{\pm4.6{\times}10^{-3}}$
% &$^{\pm8.3{\times}10^{-4}}$
% &$^{\pm4.7{\times}10^{-3}}$ \\

\midrule
\multirow[c]{8}{*}{NELL-995-re}
& \multirow[c]{2}{*}{\textbf{RelSetE}}
& \textbf{0.3198}
& \textbf{0.5611}
& \textbf{0.3996} \\
&
& $^{7.3{\times}10^{-5}}$
& $^{5.2{\times}10^{-4}}$
& $^{2.6{\times}10^{-4}}$ \\

& \multirow[c]{2}{*}{w/o ENC}
& 0.2006
& 0.3500
& 0.2509 \\
&
& $^{5.9{\times}10^{-5}}$
& $^{1.9{\times}10^{-4}}$
& $^{9.5{\times}10^{-5}}$ \\

& \multirow[c]{2}{*}{w/o AP}
& 0.2227
& 0.3951
& 0.2808 \\
&
& $^{5.0{\times}10^{-6}}$
& $^{1.5{\times}10^{-5}}$
& $^{7.6{\times}10^{-6}}$ \\

& \multirow[c]{2}{*}{w/o MHA}
& 0.1843
& 0.3279
& 0.2326 \\
&
& $^{2.9{\times}10^{-6}}$
& $^{1.8{\times}10^{-6}}$
& $^{3.2{\times}10^{-6}}$ \\

\midrule

\multirow[c]{8}{*}{NELL-1115-re}
& \multirow[c]{2}{*}{\textbf{RelSetE}}
& \textbf{0.4395}
& \textbf{0.5159}
& \textbf{0.4496} \\
&
& $^{2.9{\times}10^{-4}}$
& $^{3.3{\times}10^{-4}}$
& $^{2.8{\times}10^{-4}}$ \\

& \multirow[c]{2}{*}{w/o ENC}
& 0.2164
& 0.4916
& 0.2853 \\
&
& $^{1.0{\times}10^{-6}}$
& $^{2.5{\times}10^{-5}}$
& $^{5.0{\times}10^{-6}}$ \\

& \multirow[c]{2}{*}{w/o AP}
& 0.2108
& 0.3857
& 0.2580 \\
&
& $^{4.0{\times}10^{-6}}$
& $^{1.4{\times}10^{-5}}$
& $^{9.0{\times}10^{-6}}$ \\

& \multirow[c]{2}{*}{w/o MHA}
& 0.2851
& 0.3564
& 0.3264 \\
&
& $^{2.4{\times}10^{-4}}$
& $^{1.0{\times}10^{-3}}$
& $^{4.0{\times}10^{-4}}$ \\
\bottomrule
\end{tabular*}
\end{table}

% _________________________________NEW part 

% ------------------------End Ablation --------------
We implement three variants to study the effectiveness of the three components in RelSetE. They are: 1) w/o MHA by removing the multi-head attention layers; 2) w/o AP by replacing attention pooling with mean pooling; and 3) w/o ENC by replacing the encoder in RelSetE with a multilayer perceptron. 

As shown in Table~\ref{tab:ablation_results_vertical}, RelSetE outperforms these variants on all three datasets. All variants cause significant accuracy degradation, especially on Precision. Removing MHA causes the largest Recall drop on all three datasets (22.4\%, 41.6\%, and 30.9\%, respectively) and the largest F1 drop on FB15k-237-re and NELL-995-re, confirming its key role in capturing cross-instance dependencies. On NELL-1115-re, however, w/o AP causes the largest degradation (Precision nearly halves, from 0.4395 to 0.2108), showing attention pooling is critical for suppressing noisy instances. Replacing the encoder with an MLP (w/o ENC) also yields severe drops, e.g., F1-score falls by 37.2\% on NELL-995-re, exceeding even w/o AP on that dataset. Notably, Precision degrades more than Recall across all variants and datasets, suggesting these components mainly help filter out false positives rather than expand coverage, confirming that multi-head attention, attention pooling, and the encoder are all indispensable to RelSetE.
% -------------------END ----------------------------

\subsection{Relation Embedding Visualization}

% _____________________________________
\begin{figure*}[!htbp]
    \centering
    \includegraphics[width=0.95\linewidth]{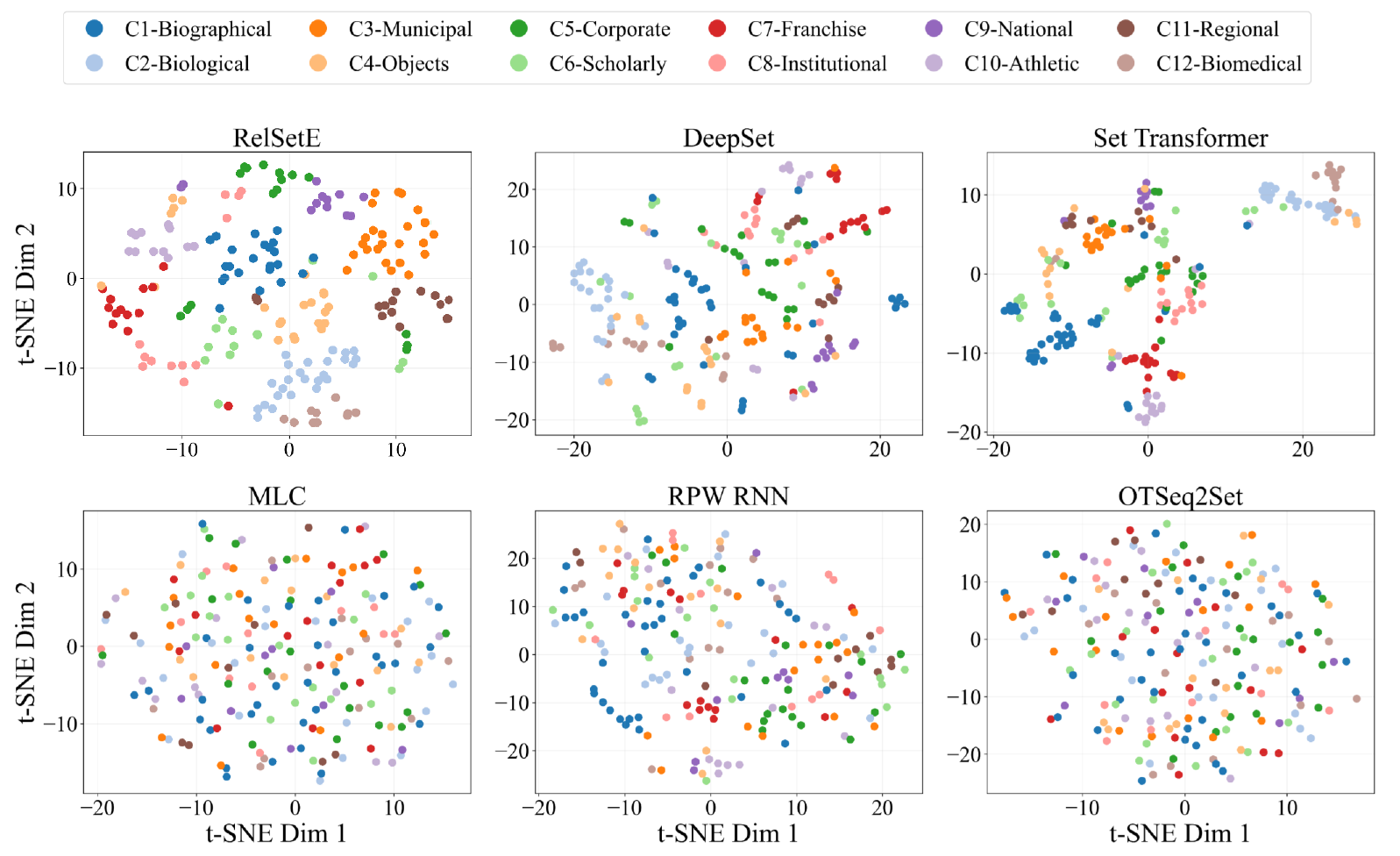}
    \caption{t-SNE visualization of relation embeddings across different models (on NELL-995-re).
    }
    \label{fig:relation_e_scatters}
\end{figure*}

\begin{figure}[!htbp]
    \centering
    \includegraphics[width=\linewidth]{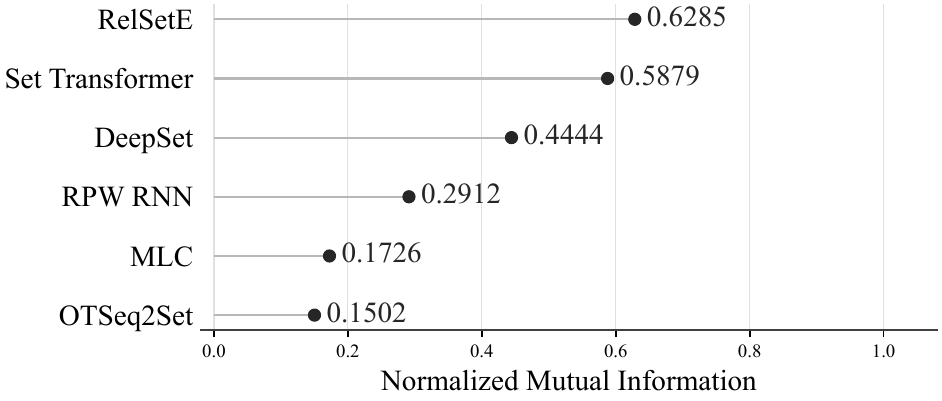}
    \caption{Relation clustering performance on NELL-995-re.}
    \label{fig:nmi}
\end{figure}

% _________________________________END________________________________
We further visualize the relation embeddings learned by RelSetE on NELL-995-re. Since the relations in NELL-995-re have descriptive names but lack predefined semantic labels, we use an LLM-assisted labeling process, followed by manual checking, to identify 12 semantically coherent categories and assign relations to them. These categories range from Biographical to Biomedical, as shown in Table~\ref{tab:sample_clusters}.
Based on these pseudo-labels, we apply t-SNE \cite{tsne} to project the relation embeddings learned by different models into a 2-dimensional space, with the results shown in Figure~\ref{fig:relation_e_scatters}.

Generally, RelSetE produces the most well-separated and compact clusters among all compared models, i.e., relations belonging to the same semantic group (e.g., C2-Biological, C9-National, C11-Regional) tend to occupy clearly distinct regions in the embedding space, with relatively sharp inter-group boundaries. DeepSet and Set Transformer show partial clustering tendencies, with some semantic groups loosely grouped, but considerable overlap remaining between others. In contrast, MLC, RPW RNN, and OTSeq2Set exhibit largely intermixed embeddings, where points from different semantic categories are scattered throughout the embedding space with little discernible cluster structure. These results suggest that RelSetE is better able to capture the underlying semantic structure of relations in its learned embeddings.

In addition, we quantitatively evaluate clustering quality by applying K-means to the relation embeddings and computing the Normalized Mutual Information (NMI) score against the pseudo-labels generated by ChatGPT. As the results shown in Figure~\ref{fig:nmi}, RelSetE achieves the highest NMI of 0.628, followed by Set Transformer (0.588) and DeepSet (0.444), while RPW RNN, MLC, and OTSeq2Set obtain considerably lower scores of 0.291, 0.173, and 0.150, respectively. This further confirms that the relation embeddings learned by RelSetE align best with the underlying semantic structure of the relations, consistent with the clustering patterns observed in the t-SNE visualization.

\begin{table}[t]
\centering
\tiny
\setlength{\tabcolsep}{2pt}
\renewcommand{\arraystretch}{0.92}
\caption{Semantic categories of relations in NELL-995-re.}
\label{tab:sample_clusters}
\begin{tabular}{p{0.4cm}p{0.9cm}p{7.2cm}}
\toprule
Category & & Relation types, shown as domain $\to$ range\\
\midrule

\clusterlabel{1} & 
Biographical & 
person $\to$ country; politician $\to$ political office; book $\to$ writer; actor $\to$ movie \\

\clusterlabel{2} &
Biological &
animal $\to$ animal; arthropod $\to$ arthropod; animal $\to$ insect; animal $\to$ invertebrate; fish $\to$ food; mammal $\to$ mammal; invertebrate $\to$ food  \\

\clusterlabel{3} &
Municipal &
river $\to$ river; television station $\to$ city; stadium/event venue $\to$ city; city $\to$ state/province; radiostation $\to$ city; transportation $\to$ city; city $\to$ geopolitical location; attraction $\to$ city  \\

\clusterlabel{4} & 
Objects &
plant $\to$ plant; clothing $\to$ clothing; visualizable thing $\to$ geometric shape; plant $\to$ emotion; sport $\to$ sportsequipment; furniture $\to$ room; visualizableobject $\to$ visualizableobject; weapon $\to$ weapon \\

\clusterlabel{5} & 
Corporate &
company $\to$ company; organization $\to$ organization; person $\to$ company; ceo $\to$ company; bank $\to$ bank \\

\clusterlabel{6} & 
Scholarly &
academic field $\to$ thing; agent $\to$ creative work; profession $\to$ tool; academic field $\to$ university; academic field $\to$ academic field \\

\clusterlabel{7} & 
Franchise &
sports game $\to$ sports team; organization $\to$ state/province; coach $\to$ sports team; sports team $\to$ city; sports team $\to$ stadium/event venue \\

\clusterlabel{8} & 
Institutional &
person $\to$ organization; organization $\to$ organization; agent $\to$ agent; television station $\to$ televisionnetwork; human agent $\to$ organization; organization $\to$ person; agent $\to$ thing \\

\clusterlabel{9} & 
National &
city $\to$ country; agricultural product $\to$ country; country $\to$ currency; country $\to$ geopolitical location; language $\to$ country \\

\clusterlabel{10} & 
Athletic &
athlete $\to$ sports team; athlete $\to$ sports team position; coach $\to$ award/trophy/tournament; athlete $\to$ sports league; athlete $\to$ stadium/event venue \\

\clusterlabel{11} & 
Regional &
mountain $\to$ state/province; 
lake $\to$ state/province; 
politician $\to$ location; 
state/province $\to$ city  \\

\clusterlabel{12} & 
Biomedical &
drug $\to$ medical condition; chemical $\to$ chemical; bacteria $\to$ medical condition \\

\bottomrule
\end{tabular}
\end{table}

\section{Conclusion}
\label{sec:conclusion}
In this work, we introduce RSC, a new task that aims to infer the set of relations semantically compatible with an entity but not yet observed in the knowledge graph, and propose RelSetE to address it, which predicts these missing relations by encoding an entity's observed relation set through a permutation-invariant encoder and a self-supervised contrastive objective. Experiments on three reconstructed benchmarks, FB15k-237-re, NELL-995-re, and NELL-1115-re, show that RelSetE consistently outperforms baselines spanning sequence-to-sequence, sequence-to-set, and set-to-set paradigms, achieving the best F1-scores across all datasets and learning relation embeddings with markedly clearer semantic structure (NMI of 0.628). Further analyses reveal that the long-tail relation distribution remains a key bottleneck, with infrequent relations being substantially harder to predict due to limited training supervision, pointing to entity neighborhood context or pre-trained language model semantics as promising directions for future work. Overall, our study establishes RSC as a practical complement to link prediction, opening up a new direction for knowledge graph completion.

\bibliographystyle{IEEEtran}
\bibliography{references}

\vfill

\end{document}